\documentclass[twoside]{article}

\usepackage{amsmath,amssymb}
\usepackage{times}
\usepackage{epsfig}
\usepackage{url}
\usepackage{graphicx,subfigure}
\usepackage[ruled, linesnumbered]{algorithm2e}
\usepackage{booktabs}
\usepackage{multirow}
\usepackage{eucal}
\usepackage{color}

\def\X{\mathbf{x}}

\def\W{\mathbf{w}}
\def\Z{\mathbf{z}}
\def\S{\mathbf{s}}
\def\P{\mathbf{p}}
\def\RR{\mathbf{r}}
\def\K{\mathbf{K}}
\def\H{\mathbf{h}}

\def\0{\mathbf{0}}

\def\XI{\boldsymbol{\xi}}
\def\PHI{\boldsymbol{\phi}}
\def\ZETA{\boldsymbol{\psi}}
\def\ALPHA{\boldsymbol{\alpha}}

\def\R{\mathbb{R}}

\def\eg{e.g.}

\newcommand{\clearemptydoublepage}{\newpage{\pagestyle{empty}\cleardoublepage}}

\begin{document}

\date{}

\title{\vspace{1cm}{\huge
Feature and Region Selection \\for Visual Learning
}}

\author{
Ji Zhao, \ Liantao Wang \\ Ricardo Cabral, \ Fernando De la Torre
}

\maketitle
\thispagestyle{empty}

\begin{center}

\vspace{-0.5cm}
{\large CMU-RI-TR-12-14}

\vspace{2cm}

{\large June 2014}

\vspace{4cm}
{\large
Robotics Institute\\
Carnegie Mellon University\\
Pittsburgh, Pennsylvania 15213\\
}

\vspace{2cm}

\copyright \/ Carnegie Mellon University

\end{center}

\pagenumbering{Roman}
\clearemptydoublepage
\setcounter{page}{1}

\begin{centering}
\section*{Abstract}
\end{centering}

{\em Visual learning problems such as object classification and action recognition are typically approached using extensions of the popular bag-of-words (BoW) model. Despite its great success, it is unclear what visual features the BoW model is learning:  Which regions in
the image or video are used to discriminate among classes? Which are the most discriminative visual words?
Answering these questions is fundamental for understanding existing BoW models and inspiring better models for visual recognition.

To answer these questions, this paper presents a method for feature selection and region selection in the  visual BoW  model. This
allows for an intermediate visualization of the features and regions that are important for visual learning. The main idea is to assign latent weights to the features or regions, and jointly optimize these latent variables with the parameters of a classifier (e.g., support vector machine).  There are four main benefits of our approach: (1) Our approach accommodates non-linear additive kernels such as the popular $\chi^2$ and intersection kernel; (2) our approach is able to handle both regions in images and spatio-temporal regions in videos in a unified way;  (3) the feature selection problem is convex, and both problems can be solved using a scalable reduced gradient method; (4) we point out strong connections with multiple kernel learning and multiple instance learning approaches. Experimental results in the PASCAL VOC 2007, MSR Action Dataset II and YouTube illustrate the benefits of our approach.
}

\clearemptydoublepage
\tableofcontents
\clearemptydoublepage

\pagenumbering{arabic}
\setcounter{page}{1}



\section{Introduction}
{T}{he} last decade has witnessed  great advances in machine learning and computer vision that have largely improved the performance and reduced the computational complexity of visual learning algorithms. Although there has been much progress  in supervised visual learning, two main limitations still exist: (1) the reliance on human labeling limits the application of supervised methods in problems involving many categories; (2) these discriminative models lack interpretability because they do not produce mid-level representations (e.g., what are the most important  visual features for discrimination?).

\begin{figure*}[!tbp]
\centering
{\includegraphics[width=1\linewidth]{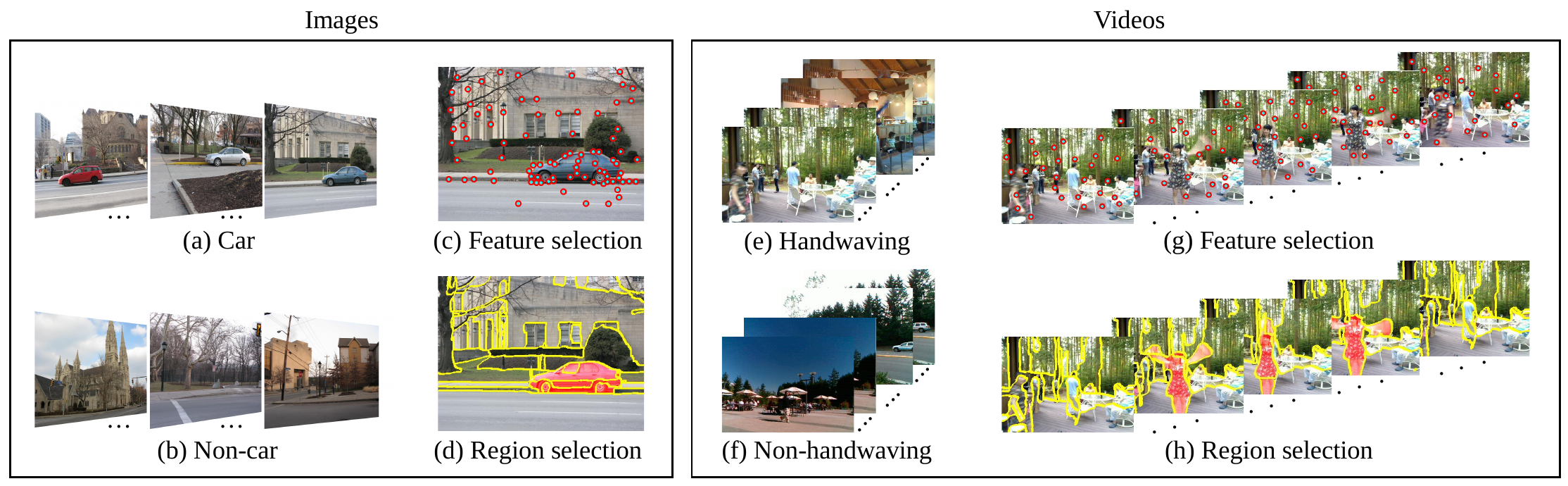}}
   \vspace{-0.2in}
\caption{ Given a set of images containing a car (a) and images without a car (b), this paper proposes an algorithm to select the visual features (c) and regions (d) that are most
discriminative in the kernel space. Similarly, given a set of videos containing hand-waving actions (e) and actions that are not hand-waving (f), we find the most
discriminative spatio-temporal features (g) and spatio-temporal regions (h). }
\label{fig:flowchart}
\end{figure*}

For instance, consider Fig.~\ref{fig:flowchart}, where there are a set of images that contain a car (Fig.~\ref{fig:flowchart} (a)) and a set of images that do not contain a car (Fig.~\ref{fig:flowchart} (b)).  Given these sets, the goal of a weakly-trained classifier is to discover discriminative regions and use them to train a car detector. Most of the successful approaches for weakly-supervised localization  (WSL)~\cite{Nguyen09,Hartmann12,Russakovsky12,YangW12,TangK13} rely on bag-of-words (BoW). BoW approaches build a vocabulary of visual words to encode the visual representation and then use it to learn a binary classifier (e.g., kernel SVM). Although these techniques achieve state-of-the-art performance, the feature spaces induced by kernels obfuscate the understanding of which are the visual features that are most important for discrimination in the image space. The aim of this paper is to develop algorithms that learn in a weakly-supervised manner which are the discriminative features and regions. We aim to answer the following questions: Which visual words are used to discriminate cars versus non-cars (Fig.~\ref{fig:flowchart}(c))? Which are the discriminative regions in the image (e.g., car in Fig.~\ref{fig:flowchart}(d))?
In addition to still images, we also apply our method to find discriminative spatio-temporal regions for activity recognition from video (Fig.~\ref{fig:flowchart} (e)-(h)).

WSL methods can partially solve the problem of localization of discriminative features, avoiding the time-consuming and error-prone manual localization process. Moreover, the  selected regions are more informative to train detectors~\cite{Nguyen09}. Due to its importance, WSL has been a popular  topic researched in the last few years. Existing algorithms for WSL rely on multiple instance learning (MIL) and have mostly been applied to linear classifiers. A major challenge is how to extend these methods to cope with kernel representations while allowing for region and feature selection, which is a non-trivial task.

This paper proposes a feature and a region selection method for visual learning in the kernel space. The feature selection method is suitable for the family of additive kernels, and the region selection is valid for all kernels.  The contributions of our work include: (1) a convex model for feature selection in the kernel space, and its application to find discriminative visual words; (2) a method for region selection using non-linear kernels, which can be used for the discovery and visualization of discriminative regions in images and spatio-temporal volumes in videos; (3) connections of our work with existing approaches including multiple kernel learning (MKL) and multiple instance learning (MIL). Experimental results in the PittCar dataset, PASCAL VOC 2007, MSR Action Dataset II and YouTube dataset illustrate the benefits of our approach.

\section{Related Work}
\subsection{Feature Selection in Kernel Space}
Selecting relevant features in kernel spaces has been a challenging problem addressed by several researchers.
Cao et al.~\cite{CaoB07} developed a feature selection method by learning feature weights in the kernel space. This procedure is done as a data processing step, independently  of the classifier construction. There also exist methods that perform feature selection and classifier construction jointly by inducing sparsity, such as \cite{Bradley98,Grandvalet02,ZhuJ04}. Here the sparsity means sparse weight, which is usually realized by imposing $L_1$ norm constraints.
We will build on previous work by Nguyen et al.~\cite{nguyen10} who proposed a convex feature weighting method for linear SVM. Our work, however, extends~\cite{nguyen10} by adding non-linear additive kernels whose effectiveness have been validated in computer vision \cite{Chatfield11,Vedaldi12}.

\subsection{Multiple Instance Learning (MIL)}
In the MIL setting, each image is modeled as a bag of regions, and each region is an instance. With two classes,  the negative bag only contains negative instances and the positive bag at least one positive. The goal of MIL is to label the positive instances within the positive bags. Many MIL algorithms have been successfully used for weakly-supervised learning, such as MILboost~\cite{Viola05}, MI-SVM~\cite{Andrews2002,Nguyen09,felzenszwalb10,YangW12} and SparseMIL \cite{Vijayanarasimhan08}.
A convex MIL method named key-instance SVM (KI-SVM) is proposed in~\cite{LiYF09}. In addition to predicting bag labels, our approach can also locate regions of interest and it has been used in content-based image retrieval.
MIL has been applied to object detection for images~\cite{felzenszwalb10,Nguyen09}, time series~\cite{Nguyen09}  and videos~\cite{Hartmann12,TangK13,Siva11}.

Among these methods, MI-SVM is arguably the most popular for WSL. However, current WSL methods based on MI-SVM have two main limitations:
(1) most approaches use bounding boxes for localization (e.g., \cite{Nguyen09,Russakovsky12}) instead of arbitrary shapes, and  (2) most methods are limited to linear kernels.
In this paper, our region selection method follows the idea of efficient region search (ERS) \cite{Vijayanarasimhan11}: an object is a certain combination of several over-segmented regions, so it can localize objects with with arbitrary shape.
Moreover, our region selection can take advantage of non-linear kernels.

\subsection{Weakly Supervised Object Localization}
Our region selection method aims to discover the discriminative regions in the positive images/videos, which turns out to be a way of weakly-supervised localization.
 In related work, Raptis et al. \cite{Raptis12} used a latent SVM to classify videos using spatio-temporal patterns.  Ghodrati et al. \cite{Ghodrati14} improved action classification by refining the recognition and video segmentation iteratively in a coupled learning framework. CRANE \cite{TangK13} modified MIL by iterating through all of the negative segments, and each negative segment penalizes nearby segment in a positive video, improving existing algorithms.  Weakly supervised localization also has a close relationship with the common pattern discovery from images that share common contents, such as co-segmentation and feature matching for sematic similar images \cite{Faktor13,MaJ16}.

There are some works that enable bag-of-words to discover informative regions automatically, which are essential for visualization and image classification.
For example, our work is most related to Liu and Wang~\cite{LiuL12}, who proposed a region of support to visualize what the BoW model has learned. However, their method uses a linear SVM and it is unclear how to extend it to the kernel domain.
Bilen et al. \cite{Bilen14} proposed a semantic representation of an object and a new latent SVM to learn the spatial location of an object for enhanced image classification. However, this method is limited to linear kernel, and depends on a careful initialization. In addition, the localization is still limited to bounding-box, while our method yields arbitrary shape, the superiority of which has been stated in \cite{Vijayanarasimhan11}.

\section{Feature Selection for Additive Kernels}
\label{sec:feat_sel}
This section proposes a convex feature selection method for additive kernels.
Let $\CMcal{S} = \{ ( \X_i, y_i ) \}^{n}_{i=1}$
(see footnote\footnote{Bold lowercase letters, such as $\P$, denote column vectors. $p_i$ represents the $i^{\text{th}}$ entry of the column vector $\P$. Non-bold letters represent scalar variables. Calligraphic uppercase letters denote sets (\eg, $\CMcal{S}$, $\CMcal{B}$).} for an explanation of the notation used in this work)
be a training set of $n$ samples, where $\X_i \in \R^D$ is the histogram of BoW for the $i^{\text{th}}$ image, $D$ is the number of visual words in the codebook, and $y_i \in \{-1, +1\}$ are the corresponding labels.

Popular choices of kernels for visual learning are additive, such as the $\chi^2$ and the histogram intersection kernels~\cite{Vedaldi12}.
Formally, a kernel $K(\cdot, \cdot)$ on $\R^D \times \R^D$ is \emph{additive} if it satisfies $ K(\X_i, \X_j) = \sum_{k=1}^D \kappa(x_{ik}, x_{jk}) $ for any samples $\X_i, \X_j \in \R^D$, where $x_{ik}$ is the $k^{\text{th}}$ bin of the BoW histogram for the $i^{\text{th}}$ image.
That is, the kernel function $\kappa(x_{ik}, x_{jk})$ is defined on one bin of the histogram.

Given an additive kernel, the goal of feature selection is to weigh the features with a weight vector in the kernel space. We parameterize the feature space with a weight vector $\P$. That is, we construct a mapping $\PHI(\X_i, \P) = [\sqrt{p_1} \ZETA^{\top}( x_{i1}), \cdots,  \sqrt{p_D} \ZETA^{\top}(x_{iD})]^{\top}$, that assigns different weights to different feature maps, where $\ZETA(x_{ik})$ is the feature map for the $k^{\text{th}}$ bin of the $i^{\text{th}}$ histogram, $\P = [p_1, \cdots, p_D]^{\top}$ are the feature weights, and $p_k \ge 0 \ \forall k$. In the maximum margin framework, we would like to find the separating hyperplane of a SVM and the feature weighting vector $\P$ that has the largest margin between classes.
However, different values of $\P$ correspond to different feature spaces, and the margins in two different feature spaces cannot be directly compared, it is necessary to normalize the margin.

\subsection{Normalized Margin SVM}
Nguyen et al.~\cite{nguyen10} defined normalized margin as the ratio of the margin $M$ over the square root of sum of squared distances (in the feature space) between same-class data instances.
Formally, the {\it normalized margin} is defined as
\begin{equation}
\frac{M}{\sqrt{\sum_{i,j=1}^n \frac{1+y_i y_j}{2} \| \PHI(\X_{i}, \P) - \PHI(\X_{j}, \P) \|^2}}.
\end{equation}
Observe that the normalized margin is invariant to scale and translation in the feature space.
The problem of finding the parameter $\P$ for the mapping and the parameters of the separating hyperplane that provides the largest normalized margin can be stated as
\begin{align}
\max_{\overline{\W}, \bar{b}, M, \P}  \ \ & \ \ \frac{M}{\sqrt{\sum_{i,j=1}^n \frac{1+y_i y_j}{2} \| \PHI(\X_{i}, \P) - \PHI(\X_{j}, \P) \|^2}} \label{nmargin} \\
 \text{s.t.} \ \ & \ \ y_i(\overline{\W}^{\top} \PHI(\X_{i}, \P) + \bar{b}) \ge M  \ \ \forall i; \nonumber \\
 & \ \| \overline{\W} \| = 1. \nonumber
\end{align}
We can see that if $\P$ is fixed, finding the hyperplane with the maximum normalized margin is equivalent to finding the hyperplane that maximizes the normal margin $M$.

Let $\W = \overline{\W}/M$, $b = \overline{b}/M$, and denote the normalization factor
\begin{align}
\varphi(\P) = \sum_{i,j=1}^n \frac{1+y_i y_j}{2} \| \PHI(\X_{i}, \P) - \PHI(\X_{j}, \P) \|^2. \label{equ:nmlz}
\end{align}
Then $M=\|\overline{\W}\|/\|\W\|=1/\|\W\|$. Substituting $\varphi(\P)$ into problem~\eqref{nmargin}, we obtain an equivalent problem
\begin{align*}
\max_{\W, b, \P}  \ \ & \ \ \frac{1}{\sqrt{\varphi(\P)} \ \|\W\| }\\
 \text{s.t.} \ \ & \ \ y_i(\W^{\top} \PHI(\X_{i}, \P) + b) \ge 1  \ \ \forall i. \nonumber
\end{align*}
The above problem is again equivalent to
\begin{align*}
\min_{\W, b, \P}  \ \ & \ \ \frac{1}{2} \varphi(\P) \| \W \|^2 \\
 \text{s.t.} \ \ & \ \ y_i(\W^{\top} \PHI(\X_{i}, \P) + b) \ge 1  \ \ \forall i. \nonumber
\end{align*}
Using soft-margin instead of hard-margin, the above formulation can be converted to
\begin{align}
\min_{\W, b, \P, \XI}  \ \ & \ \ \frac{1}{2} \varphi(\P) \| \W \|^2 + C \sum_{i=1}^n \xi_i \label{nmargin_slack} \\
 \text{s.t.} \ \ & \ \ y_i(\W^{\top} \PHI(\X_{i}, \P) + b) \ge 1 -\xi_i  \ \ \forall i; \ \ \XI \ge \0. \nonumber
\end{align}
Here, $\XI = [\xi_1, \cdots, \xi_n]$ are slack variables which allow for penalized constraint violation, and $C$ is the parameter that controls the trade-off between generalization and training error.

\subsection{Normalized Margin SVM with Additive Kernels}
In \cite{nguyen10}, they only solve the SVM with normalized margin for linear kernels.
In this paper, we propose a method to solve the SVM with normalized margin for additive kernels in problem~\eqref{nmargin_slack}.

In order to transform problem~\eqref{nmargin_slack} into a convex optimization problem and solve it efficiently, we make use of two properties of additive kernels. First, as we have mentioned, $\PHI(\X_i, \P) = [\sqrt{p_1} \ZETA^{\top}( x_{i1}), \cdots,  \sqrt{p_D} \ZETA^{\top}(x_{iD})]^{\top}$ for additive kernel, so the normalization factor $\varphi(\P)$ in Eq.~\eqref{equ:nmlz} can be re-written as
\begin{align}
\varphi(\P) = \sum_{k=1}^D a_k p_k,
\end{align}
where
\begin{align}
a_k \ &= \sum_{i,j=1}^n \frac{1+y_i y_j}{2} \| \ZETA(x_{ik}) - \ZETA(x_{jk}) \|^2  \label{equ:ak} \\
& =  \sum_{i,j=1}^n \frac{1+y_i y_j}{2} \left[ \kappa(x_{ik}, x_{ik}) - 2\kappa(x_{ik}, x_{jk}) +  \kappa(x_{jk}, x_{jk})\right].
\nonumber
\end{align}
Note that  $a_k$
can be interpreted as the total distance of the $k^\text{th}$ bin in kernel space, and it can be computed from the training data a priori.
Other normalization factors can also be utilized without additional innovation. In \cite{Feng15}, it provides a rather encyclopedic list of alternatives.

Second, the hyperplane $\W$ can be re-written as a vertical concatenation of  $D$ column vectors as $\W = [\W_1^{\top}, \cdots,  \W_D^{\top}]^{\top}$, where each $\W_k$ weighs the feature map for each bin $\ZETA(x_{ik})$. Then the following two equations hold: $\W^{\top} \PHI(\X_i, \P) = \sum_{k=1}^D \sqrt{p_k} \W_k^{\top} \ZETA(x_{ik})$, and $\| \W \|^2 = \sum_{k=1}^D \| \W_k \|^2$.

Since $\varphi(\P)$ is homogeneous in $\P$, we can always scale $\P$ appropriately to
get $\varphi(\P) = 1$. Using this constraint, and making a variable substitution $\W_k \leftarrow \sqrt{p_k}\W_k$, problem~\eqref{nmargin_slack} can be written as
\begin{align}
\min_{\W, b, \P, \XI} \ \ & \ \frac{1}{2} \sum_{k=1}^D \frac{\|\W_k\|^2}{p_k}  + C \sum_{i=1}^n \xi_i \label{OpGeneral2} \\
\text{s.t.}  \ \ & \ \ y_i \left[ \sum_{k=1}^D \W_k^{\top}  \ZETA(x_{ik})  + b \right] \ge 1 - \xi_i \ \forall i;  \nonumber \\ & \ \ \sum_{k=1}^D a_k p_k = 1; \ \  \P \ge \0; \ \  \XI \ge \0. \nonumber
\end{align}
where we use the convention that $\frac{t}{0}=0$ if $t=0$ and $\infty$ otherwise.
Problem~\eqref{OpGeneral2} is convex, and we propose a scalable optimization strategy in Section~\ref{sec:feat_selec_opt}.

\subsection{Relation to Multiple Kernel Learning}
We note the remarkable relationship between our feature selection formulation in problem~\eqref{OpGeneral2} and multiple kernel learning (MKL)~\cite{Rakotomamonjy08,Gehler08,Kloft11}, with the main difference being the constraints on $\P$.
In MKL, the constraint is that $\P$ lies on the probability simplex, i.e., $\sum_{k=1}^D p_k = 1$ and $p_k>0 \ \forall k$.
In our feature selection formulation, the constraint is data-driven and adaptive, i.e., $\sum_{k=1}^D a_k p_k = 1$ and $p_k>0 \ \forall k$.
Weighing each bin differently will result in increased accuracy because normalized margin SVM is expected to assign higher weights to more informative bins. Besides, feature weighting can avoid the mis-domination of the bins with larger numeric ranges to those with smaller numeric ranges.

Our feature selection method used a normalized SVM margin for feature selection with additive kernels. By leveraging the properties of additive kernels, the normalized SVM margin  is converted to a MKL alike problem. As a result, problem~\eqref{OpGeneral2} can also be interpreted as a MKL with normalized margin to handle the feature scaling problem. There are some works that incorporate the radius of minimum enclosing ball (MEB) into MKL to address kernel scaling issue \cite{Do09,GaiK10}. Liu et al. \cite{Liu12} incorporated the radius information in a more robust and efficient way to avoid complex learning structure and high computational cost.

\subsection{Optimization with the Reduced Gradient Method}
\label{sec:feat_selec_opt}
The connection between our feature selection method and MKL allows us to exploit the existing algorithms for MKL.
We can derive a scalable algorithm with proven convergence properties by optimizing problem~\eqref{OpGeneral2} with a reduced gradient
method~\cite{Rakotomamonjy08}.
For fixed $\W, b, \XI$, problem~\eqref{OpGeneral2} can be reformulated as a non-linear objective function with constraints over the simplex on $\P$. Formally,
\begin{align}
\min_{\P} J(\P) \text{  such that  } \sum_{k=1}^D a_k p_k = 1, \ p_k \ge 0,
\label{OpGeneralJP}
\end{align}
where
\begin{align}
J(\P) = \left\{
\begin{array}{rl}
\displaystyle{\min_{\W, b, \XI}} & \displaystyle{\frac{1}{2} \sum_{k=1}^D \frac{\|\W_k\|^2}{p_k}  + C \sum_{i=1}^n \xi_i} \\
\text{s.t.} & \displaystyle{y_i \left[ \sum_{k=1}^D \W_k^{\top}  \ZETA(\X_{ik})  + b \right] \ge 1 - \xi_i; \forall i} \\
 & \XI \ge \0.
\end{array} \right.
\label{OpGeneral3}
\end{align}
To use a reduced gradient algorithm to optimize this problem, we first computed the gradient $\frac{\partial J}{\partial \P}$ and then calculate reduced gradient $\nabla_{\text{red}}J$ and descent direction $\RR$ based on the gradient and constraints on $\P$.

To solve the problem, we introduced Lagrange multipliers $\alpha_i$ and $\beta_i$ for the first and second constraints in problem~\eqref{OpGeneral3}, respectively.
By setting the derivatives of the Lagrangian of problem~\eqref{OpGeneral3}  with respect to the primal variables $\W$, $b$, $\XI$ to zero,
we get the associated dual problem
\begin{align}
\max_{\ALPHA} \ & \ -\frac{1}{2} \sum_{i,j=1}^n \alpha_i \alpha_j y_i y_j \sum_{k=1}^D p_k \kappa(x_{ik}, x_{jk}) + \sum_{i=1}^n \alpha_i
\label{equ:alpha} \\
\text{s.t.} \ & \ \sum_{i=1}^n \alpha_i y_i = 0;
 \quad 0 \le \alpha_i \le C \  \forall i. \nonumber
\end{align}
This dual problem is identified as the standard SVM dual problem using the combined kernel $K(\X_i, \X_j) = \sum_{k=1}^D p_k \kappa(x_{ik}, x_{jk})$. Because of strong duality, the objective value of this dual problem~\eqref{equ:alpha} is also $J(\P)$.
Existence and computation of derivatives of $J(\P)$ have been discussed in previous literature~\cite{Rakotomamonjy08}.
Taking advantage of these previous works, the differentiation of the dual function with respect to $p_k$ is
\begin{align}
\frac{\partial J}{\partial p_k} = -\frac{1}{2} \sum_{i,j=1}^n \alpha_i^{*} \alpha_j^{*} y_i y_j \kappa(x_{ik}, x_{jk}) \ \forall k,
\label{equ:partialJ1}
\end{align}
where $\alpha^*$ maximizes the objective function in problem~\eqref{equ:alpha}.

Once the gradient of $J(\P)$ is computed, $\P$ is updated using a descent direction ensuring that the equality constraint and the non-negativity constraints on $\P$ are satisfied. Let $p_{\mu}$ be the largest entry of $\P$. The reduced gradient of $J(\P)$, denoted $\nabla_{\text{red}}J$, can be written as
\begin{align}
[\nabla_{\text{red}}J]_k = \left\{
\begin{array}{ll}
\displaystyle{\frac{\partial J}{\partial p_k} - \frac{a_k}{a_\mu} \frac{\partial J}{\partial p_\mu}} &  \text{if } k\neq\mu; \\
\displaystyle{\sum_{v\neq\mu} \left( \frac{a_v^2}{a_{\mu}^2} \frac{\partial J}{\partial p_\mu} - \frac{a_v}{a_{\mu}} \frac{\partial J}{\partial p_v} \right)} & \text{if } k = \mu.
\end{array} \right. \label{equ:nabla}
\end{align}
Descent direction is in the opposite direction with reduced gradient. However, the positivity constraints should be taken into account in the descent direction.
If $p_k=0$ and $[\nabla_{\text{red}}J]_k >0$, using this descent direction would violate the positivity constraint for $p_k$. Thus, the descent direction for that component should be set to 0. Therefore, the descent direction for updating $\P$ is
\begin{align}
\RR_k = \left\{
\begin{array}{ll}
0 \qquad \quad \displaystyle{\text{if } p_k = 0 \text{ and } \frac{\partial J}{\partial p_k} - \frac{a_k}{a_\mu} \frac{\partial J}{\partial p_\mu} >0;} \\
\displaystyle{-\frac{\partial J}{\partial p_k} + \frac{a_k}{a_\mu} \frac{\partial J}{\partial p_\mu}} \qquad  \text{if } p_k>0 \text{ and } k\neq\mu; \\
\displaystyle{\sum_{v\neq\mu, p_v>0} \left( -\frac{a_v^2}{a_{\mu}^2} \frac{\partial J}{\partial p_\mu} + \frac{a_v}{a_{\mu}} \frac{\partial J}{\partial p_v} \right)} \ \text{if } k = \mu.
\end{array} \right. \label{OpGeneral4}
\end{align}

The usual updating scheme is $\P \leftarrow \P + \gamma \RR$, where $\gamma$ is the step size. $\gamma$  is calculated using a line search method. For each $\gamma$ during the line search, we obtained a new $\P$ and used an SVM solver to calculate problem~\eqref{equ:alpha}.

We summarize the training of feature selection with additive kernels in Algorithm~\ref{alg:algo_fs}.
For testing, the prediction function is
\begin{align*}
f(\Z) & = \sum_{i=1}^n y_i \alpha_i K(\Z, \X_i) + b = \sum_{i=1}^n y_i \alpha_i \sum_{k=1}^D  p_k \kappa(z_k, x_{ik})  + b.
\end{align*}

\begin{algorithm}[t]
\caption{Feature Selection: Normalized Margin SVM with Additive Kernels} \label{alg:algo_fs}
\KwIn{Training set $\CMcal{S} = \{ ( \X_i, y_i ) \}^{n}_{i=1}$; kernel for each bin $\kappa(x_{ik}, x_{jk}) = \langle \ZETA(x_{ik}), \ZETA(x_{jk}) \rangle$; penalty coefficient $C$.}
\KwOut{Weight $\P$; SVM parameters $\ALPHA$ and $b$.}
Initialize $p_{k} = 1/\sum_{i=1}^D a_i, \forall k$\;
Calculate $a_k$ by Eq.~\eqref{equ:ak}\;
\While{stopping criterion not met}{
  Solve problem~\eqref{equ:alpha} by an SVM solver to update $\ALPHA$ and $b$\;
  Calculate $\frac{\partial J}{\partial \P}$ by Eq.~\eqref{equ:partialJ1}\;
  Set $\mu = \arg\!\max_{\mu} \P_{\mu}$\;
  Calculate the descent direction $\RR$ by Eq.~\eqref{equ:nabla}\eqref{OpGeneral4}\;
  Line search along $\RR$ to find the optimal step $\gamma$\;
  Update $\P \leftarrow \P + \gamma \RR$\;
}
\end{algorithm}

\section{Region Selection for Weakly Supervised Visual Learning}
\label{sec:reg_sel}
In the previous section, we have proposed a feature selection method in the kernel space for additive kernels. However,
visual features are typically very sparse and it is difficult to assess which regions the classifier uses for learning. In this section, we propose a method for selecting discriminative regions in images and videos. Prior to applying our method, we over-segment the images and videos into regions, i.e. superpixels~\cite{Arbelaez11}  or  spatio-temporal regions~\cite{ChenYC12}. Once the regions are segmented, we encoded each region using the BoW codebook learned from all training images/videos.
We assumed an additive property of the classifier for region selection so that the classifier score of an image is a weighted sum of the score for each of the regions.

\subsection{Weakly Supervised Localization as Region Selection}
Given an over-segmentation for each image (or  video) into $m_i$ regions, $\H_{ik}$ and $s_{ik}$  represent the BoW histogram and the importance (weight)
for the $k^{\text{th}}$ region in the $i^{\text{th}}$ image.  Our SVM for region
selection minimizes
\begin{align}
\min_{\W, b, \{\S_i\}, \XI} \ \ & \ \frac{1}{2} \|\W\|^2  + C_1 \sum_{i \in \CMcal{B}^+} \xi_i^+ + C_2 \sum_{i \in \CMcal{B}^-} \sum_{k=1}^{m_i} \xi_{ik}^- \label{equ:OpRegion1} \\
\text{s.t.} \ \ & \ \sum_{k=1}^{m_i} s_{ik} \W^{\top} \PHI(\H_{ik})  + b \ge 1 - \xi_i^+ \ \forall i \in \CMcal{B}^+; \nonumber \\
 & \ \ - \W^{\top} \PHI(\H_{ik}) - b \ge 1 - \xi_{ik}^- \ \forall i \in \CMcal{B}^-, \quad \forall k \in \{1, \cdots, m_i\}; \nonumber \\
 & \ \ \| \S_{i} \|_1 = 1, \S_i \ge \0 \ \forall i \in \CMcal{B}^+; \ \XI \ge \0. \nonumber
\end{align}
where $\PHI(\cdot)$ is the kernel feature map. $\CMcal{B}^+$ and $\CMcal{B}^-$ are index sets of training samples with label $+1$ and $-1$, respectively.
$C_1$ and $C_2$ trade-off the model complexity and empirical losses on the positive and negative bags, respectively.
The first constraint is imposed on the positive bags, and enforces that, for positive images, a combination of its segments' scores is expected to be positive or it will be penalized.
The second constraint enforces that all the segments' scores of the negative images should be negative.
The third constraint enforces that $\S_i$ lies on the probability simplex. Thus the solution tends to be sparse and can be used for region selection\footnote{In the paper, we refer
to region as a set of superpixels in images or spatio-temporal regions in videos. The problem of~\eqref{equ:OpRegion1} is one of region weighting. We call it region selection since the solution is sparse and only a few regions have non-zero weights.}.
If we impose $L_q$ norm constraint with $q>1$ on $\S_i$, it will generate non-sparse solutions \cite{Kloft11}.

{\bf Prediction}
Once the SVM parameters are learned, the classification and localization for new test images can be performed simultaneously. Given
the $i^{\text{th}}$ image and its over-segmented regions (indexed by $k$), we can provide an initial estimate if a region belongs to a discriminative region or not by computing
the decision value $\W^{\top} \PHI(\H_{ik}) + b$.  The final score of the image is the weighed average score of its regions, that is,  $\sum_{k} s_{ik} \W^{\top} \PHI(\H_{ik})  + b$. The weights $s_{ik}$ are learned during training.

\subsection{Relation to Multiple Instance Learning}
The proposed region selection method has closed connection to multiple instance learning (MIL) algorithms. MIL makes the assumption that a negative bag contains only  negative instances, whereas and a positive bag has at least one positive instance.  However, in our region selection method, the bag label is determined by a combination of regions. This is a more reasonable assumption for visual learning because it is difficult to determine which region triggers a label for an image, considering that the segmentation may not yield perfect results. Generally speaking, in MIL, the label is determined by the maximum of the instances scores, while in our method, the label is determined by the weighted mean of all the instances' scores.

Our formulation is different from previous key-instance SVM (KI-SVM), where it is assumed  that there is only one positive instance in each positive bag~\cite{LiYF09}. Our formulation is also different from kernel latent SVM (KLSVM)~\cite{YangW12}, which also relies on a single instance to determine the label for positive bags. In \cite{Yakhnenko11}, the method scores an image using the combination of regions, but it is limited to the linear kernel case. Note that our region selection method in this section is compatible with any kernel.

\subsection{Optimization with the Reduced Gradient Method}
\label{sec:reg_sel_opt}
Similar to the feature selection problem~\eqref{OpGeneral2}, the region selection problem~\eqref{equ:OpRegion1} can also be reformulated as a non-linear objective function with constraints
over the simplex. We used the reduced gradient  method to solve it with a coordinate descent strategy.
First, we fixed the weights $\S$, and optimized the object function w.r.t. $\W$, $b$ and $\XI$. Second, we used the reduced gradient method to update $\S$.

In order to simplify the notation, we took each region in a negative image as a negative bag that contains only one instance.  We set $C_2$ equal to $C_1$, and reformulate problem~\eqref{equ:OpRegion1} as
\begin{align}
\min_{\{\S_i\}} J(\{\S_i\}) \text{  such that  } \| \S_{i} \|_1 = 1, \ \S_i \ge \0 \ \forall i,
\end{align}
where
\begin{align}
J(\{\S_i\}) = \left\{
\begin{array}{rl}
\displaystyle{\min_{\W, b, \XI}} & \displaystyle{\frac{1}{2} \|\W\|^2  + C \sum_{i=1}^n \xi_i} \\
\text{s.t.} & \displaystyle{y_i \left[ \W^{\top} \sum_{k=1}^{m_i} s_{ik} \PHI(\H_{ik})  + b \right] \ge 1 - \xi_i \  \forall i} \\
 & \XI \ge \0.
\end{array} \right.
\label{equ:OpRegion2}
\end{align}
By setting the derivatives of the Lagrangian of problem~\eqref{equ:OpRegion2} to zero, we get the associated dual problem
\begin{align}
\max_{\ALPHA} \ & \ -\frac{1}{2} \sum_{i,j=1}^n \alpha_i \alpha_j y_i y_j \left( \sum_{k=1}^{m_i} \sum_{l=1}^{m_j} s_{ik} s_{jl} K(\H_{ik}, \H_{jl})  \right) + \sum_{i=1}^n \alpha_i  \label{equ:alpha2} \\
\text{s.t.} \ & \  \sum_{i=1}^n \alpha_i y_i = 0; \ \ 0 \le \alpha_i \le C \  \forall i. \nonumber
\end{align}
This is the standard dual formulation for SVM with the combined kernel $K(\H_i, \H_j) = \sum_{k=1}^{m_i} \sum_{l=1}^{m_j} s_{ik} s_{jl} K(\H_{ik}, \H_{jl})$. Because of strong duality, $J(\{\S_i\})$ is also the objective value of this dual problem. By differentiating the dual function with respect to $s_{ik}$, we have
\begin{align}
\frac{\partial J}{\partial s_{ik}} = -\frac{1}{2} \sum_{j=1}^n \alpha_i^{*} \alpha_j^{*} y_i y_j \sum_{l=1}^{m_j} s_{jl} K(\H_{ik}, \H_{jl}), \label{equ:partialJ2}
\end{align}
where $\alpha^*$ maximizes problem~\eqref{equ:alpha2}.
After the gradient $\frac{\partial J}{\partial s_{ik}}$ has been calculated, we can get the reduced gradient and descent direction using the way in Section~\ref{sec:feat_selec_opt}.

At first glance, computing the gradient in Eq.~\eqref{equ:partialJ2} seems to be computationally expensive. However, this calculation is efficient for the following reasons. First, we can reformulate it as a compact matrix formulation when calculating $\frac{\partial J}{\partial \S_{i}}$. Second, since  ${\bf \alpha}$ is sparse, the complexity of calculating gradient is largely reduced. The region selection method is sumarized in Table of Algorithm~\ref{alg:algo1}.

\begin{algorithm}[t]
\caption{Region Selection Algorithm} \label{alg:algo1}
\KwIn{Training set $( \{(\X_{ik})\}_{k=1}^{m_i}, y_i)_i$, $i=1, \cdots, n$; kernel $K(\cdot, \cdot)$; penalty coefficient $C$.}
\KwOut{Region annotation $\{(s_{ik})\}_{k=1}^{m_i}$, $i=1, \cdots, n$; SVM parameters $\ALPHA$ and $b$.}
Initialize $s_{ik} = 1/m_i, \forall i, k$\;
Construct block kernel matrix $\widetilde{\K}$. The $(k,l)$-th element of the $(i,j)$-th block is defined as $[\widetilde{\K}(i,j)]_{kl} = K(\X_{ik}, \X_{jl})$\;
\While{stopping criterion not met}{
  Calculate kernel matrix $\K$ with its element $\K_{ij} = \S_i^{\top} \widetilde{\K}(i,j) \S_j$\;
  Calculate $J(\{\S_i\})$ in \eqref{equ:OpRegion2} by an SVM solver with kernel matrix $\K$, get SVM parameters $\ALPHA$ and $b$\;
  Calculate $\frac{\partial J}{\partial \S_{i}}$ for $i = 1, \cdots, n$ by Eq.~\eqref{equ:partialJ2}\;
  Calculate reduced gradient and descent direction $\RR_i, \forall i$\;
  Line search to find optimal step $\gamma_i$ for $\S_i \ \forall i$\;
  Update $\S_i \leftarrow \S_i + \gamma_i \RR_i$, $i = 1, \cdots, n$\;
}
\end{algorithm}


\section{Experimental Results}
\begin{figure*}[!tb]
  \centering
  {\includegraphics[width=1\linewidth]{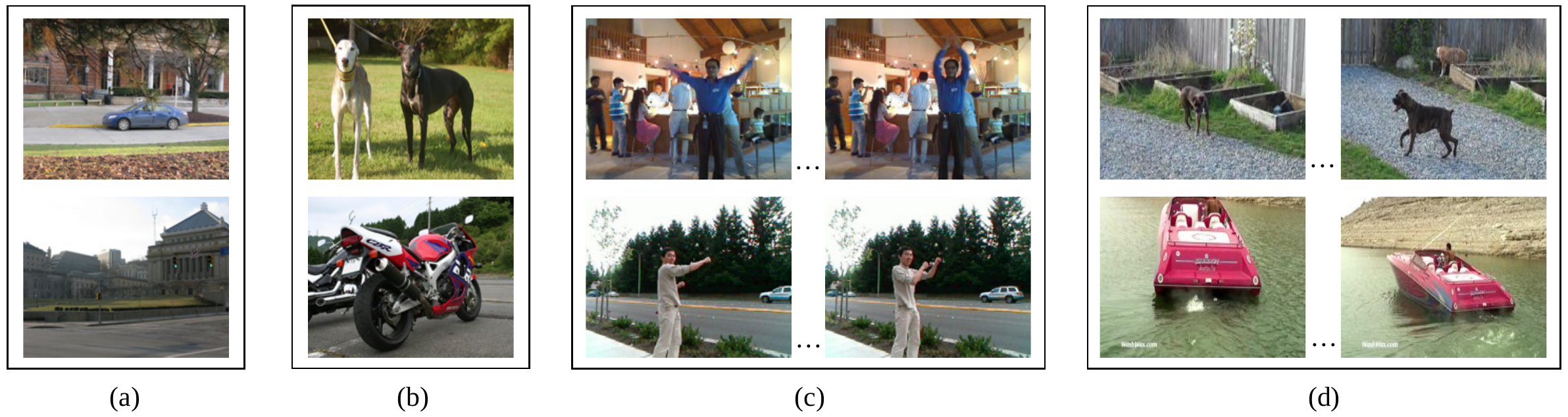}}
     \vspace{-0.2in}
  \caption{Some examples of the datasets.  (a) PittCar; (b) PASCAL VOC; (c) MSR Action II; (d) YouTube Objects.}
  \label{fig:dataset}
\end{figure*}

This section validated the performance of our feature selection and region selection algorithms by comparing them with other state-of-the-art approaches on the following four datasets:

\noindent\textbf{PittCar Dataset}~\cite{Nguyen09} contains $400$ images of which 200 are positive and 200 negative, see Fig.~\ref{fig:dataset}a. There is only one object in each positive image. Half of the positive and negative images were used as training data, and the rest were used for testing. For each image, we extracted SIFT features~\cite{lowe04} densely and selected $10000$ of them randomly. All the SIFT descriptors were quantized into $1000$ visual words, obtained by applying K-means to $100000$ training samples.

\noindent\textbf{PASCAL VOC 2007} consists of $9963$ images. For examples see Fig.~\ref{fig:dataset}b. There are $20$ object categories, with some images containing multiple objects. This dataset has been previously split into training and testing sets, which contained $5011$ and $4952$ images respectively. We proceeded as in the PittCar Dataset, extracting SIFT features and building a codebook of $1000$ dimensions.

\noindent\textbf{MSR Action Dataset II}~\cite{YuanJ11} comprises $54$ video sequences of crowded environments, see Fig.~\ref{fig:dataset}c. There are $3$ action categories: hand waving, handclapping, and boxing. Each video sequence contains multiple actions.
Following~\cite{Siva11}, we split each video to contain only one action and randomly selected $135$ videos as training data and $46$ for test data. During this random division, the videos containing multiple actions that could not be split temporally were always included in the testing set. We extracted STIP features~\cite{Laptev05} densely for each video. All the feature points were then quantized into 2000 words, which were obtained by applying K-means to $100000$ training descriptors.

\noindent\textbf{YouTube-Objects (YTO)} \cite{Prest12} consists of videos collected from YouTube, see Fig.~\ref{fig:dataset}d. It contains $10$ of the $20$ classes in the PASCAL VOC. Tang {\it et al} ~\cite{TangK13} generated a ground truth set of $151$ shots by manually annotating segments after the segmentation.  We used the features in~\cite{TangK13} that include histograms of dense-SIFT, histograms of RGB color, histograms of local binary patterns, histograms of dense optical flow, and heat maps.

\subsection{Feature Selection Experiments} \label{sec:exp_feat_sel}

To validate the effectiveness of the proposed feature selection method, we compared our feature selection with $\chi^2$ kernel with the following baselines: (i) Linear SVM; (ii) $\chi^2$ kernel SVM; (iii) feature selection with linear SVM \cite{nguyen10}; (iv)   MKL using $\chi^2$ kernel \cite{Rakotomamonjy08}, due to their connection with our method explained in Section~\ref{sec:feat_sel}. For MKL, each kernel is defined on one bin of the histograms.

For each method, parameters (e.g., C in the SVM) were chosen via cross-validation and we measured the classification performance using average precision (AP). To assess the complexity reduction achieved by feature selection, we also measured the number of selected features (i.e., non-zero weight). In this case, the features are the bins (clusters) in the BoW model.
The results are presented in Table~\ref{tab:car} for PittCar and MSR II datasets. We can see that our feature selection with $\chi^2$ kernel (FS-$\chi^2$) achieved the best average precision (AP) in all cases except `Boxing', where it is outperformed only by the linear kernel, while the number of our selected features is significantly smaller than the original feature dimension.
In Table~\ref{tab:pascal} for PASCAL VOC 2007 dataset, the feature selection for $\chi^2$ kernel SVM achieved comparable mean AP than $\chi^2$ SVM ($0.373$ vs $0.375$) over 20 classes, but used much less features ($265$ vs $1000$).

\begin{figure*}[!tb]
  \centering
  {\includegraphics[width=1\linewidth]{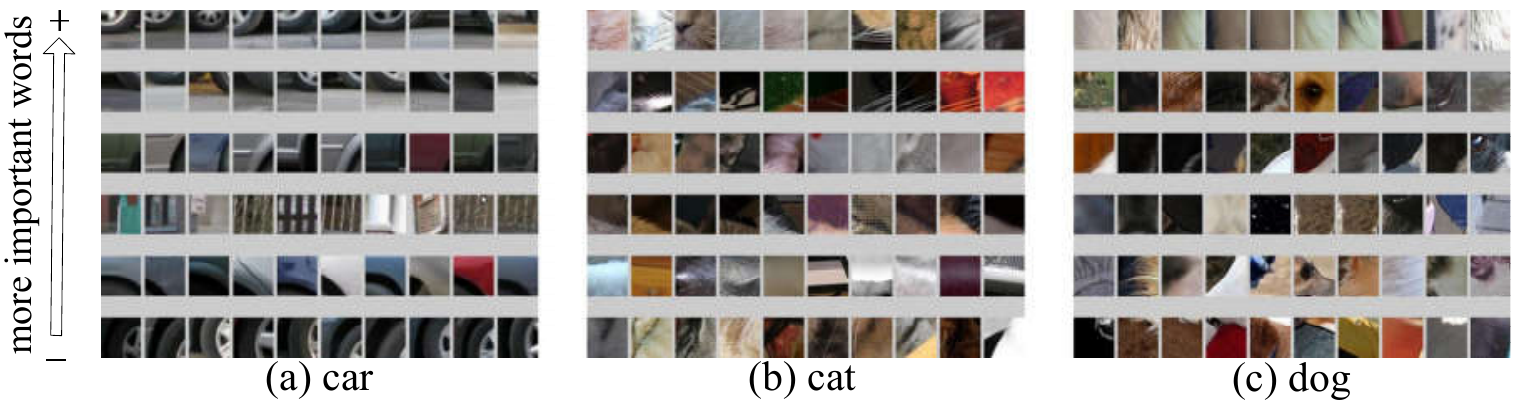}} \
    \vspace{-0.2in}
  \caption{Patch visualization of top $6$ visual words with highest weights in the feature selection. (a) Car in PittCar dataset, (b) Cat in PASCAL VOC 2007, (c) Dog in PASCAL VOC 2007. Each row line has $10$ randomly selected patches corresponding to the visual word. From top to bottom, the weight changes from high to low.
}
\label{fig:dict}
\end{figure*}

A major goal of the paper is to illustrate that by performing feature and region selection, we can achieve a better interpretability of the BoW model.
We visualized the selected visual words in the codebook for PittCar and PASCAL VOC 2007 datasets, in Fig.~\ref{fig:dict}. From the feature selection results on the PittCar dataset, we can see that the most discriminative features mainly come from the wheels and doors of the cars. Note that the visual word with the fourth largest weight corresponds to the trunks of trees and fences.
This is because trees occur more frequently in negative images than in positive images. As a result, this visual word is selected as a discriminative. For the cat and dog classes in PASCAL VOC dataset, several words latch on to cat and dog faces, while other visual words represent context (e.g.,  carpets) in which these animals usually appear. Since our method allows us to visualize the patches of visual words with their weights, the irrelevant words can be easily interpret by looking at the images in the dataset. From this example, we can see that feature selection can reveal which context the classifier is using for discriminating among classes.

\begin{table*}[!tb]
\small
\centering
\caption{The comparison of classification performance for feature selection \protect \\ methods and MKL on the PittCar and MSR Action II datasets.}
\label{tab:car}
\begin{tabular}{|l|c|r|c|r|c|r|c|r|}
\hline
\multirow{3}{*}{} & \multicolumn{2}{c|}{\multirow{2}{*}{PittCar}} & \multicolumn{6}{c|}{MSR Action II}                                                 \\ \cline{4-9}
                  & \multicolumn{2}{c|}{} & \multicolumn{2}{c|}{Hand Clapping} & \multicolumn{2}{c|}{Hand Waving} & \multicolumn{2}{c|}{Boxing} \\ \cline{2-9}
                  & AP & \#Feat & AP & \#Feat & AP & \#Feat & AP & \#Feat \\ \hline
linear SVM  & 0.833 & 1000 & 0.528 & 2000 & 0.630 & 2000 & 0.716 & 2000 \\ 
$\chi^2$ SVM & 0.959 & 1000 & 0.563 & 2000 & 0.699 & 2000 & 0.680  & 2000 \\ 
MKL-$\chi^2$~\cite{Rakotomamonjy08} & 0.961 & 120 & 0.687 & 102 & 0.741 & 96 & 0.810 & 112 \\ 
FS-linear~\cite{nguyen10} & 0.967 & 112 & {\bf 0.717} & 72 & 0.832 & 87 & {\bf 0.897} & 83 \\ \hline
FS-$\chi^2$ (ours) & {\bf 0.988} & 56 & {\bf 0.717} & 79 & {\bf 0.847} & 56 & 0.852 & 45 \\ \hline
\end{tabular}
\end{table*}

\begin{table*}[!tb]
\small
\centering
\caption{The comparison of classification performance for feature selection \protect \\ methods and MKL on the PittCar and MSR Action II datasets.}
\label{tab:pascal}
\begin{tabular}{|l|c|r|c|r|c|r|c|r|c|r|}
\hline
\multirow{2}{*}{} & \multicolumn{2}{c|}{aeroplane} & \multicolumn{2}{c|}{bicycle} & \multicolumn{2}{c|}{bird} & \multicolumn{2}{c|}{boat} & \multicolumn{2}{c|}{bottle}  \\ \cline{2-11}
                  & AP & \#Feat & AP & \#Feat & AP & \#Feat & AP & \#Feat & AP & \#Feat \\ \hline
linear SVM & 0.501 & 1000 & 0.274 & 1000 & 0.255 & 1000 & 0.418 & 1000 & 0.120 & 1000\\ 
$\chi^2$ SVM & 0.516    &   1000    &   0.384   & 1000      &  {\bf 0.295}    &  1000     &  {\bf 0.456}  & 1000      & 0.193     & 1000    \\ 
MKL-$\chi^2$ &   0.484   &  68       & 0.356     & 56        & 0.280     & 691       & 0.416     & 351       &  0.190  & 675      \\ 
FS-linear &   0.392   & 364       &   0.314   &  397      &  0.241    & 661       & 0.323     & 358       &  0.147    &  396        \\ \hline
FS-$\chi^2$  &   {\bf 0.517}    & 63    & {\bf 0.397} &  54   & 0.277   &  690    & 0.443   & 67     & {\bf 0.198}& 491    \\ \hline
\end{tabular}
\newline
\vspace*{0.1in}
\newline
\begin{tabular}{|l|c|r|c|r|c|r|c|r|c|r|}
\hline
\multirow{2}{*}{} & \multicolumn{2}{c|}{bus} & \multicolumn{2}{c|}{car}  & \multicolumn{2}{c|}{cat} & \multicolumn{2}{c|}{chair} & \multicolumn{2}{c|}{cow} \\ \cline{2-11}
                  & AP & \#Feat & AP & \#Feat & AP & \#Feat & AP & \#Feat & AP & \#Feat  \\ \hline
linear SVM  & 0.249 & 1000 & 0.468 & 1000 & 0.290 & 1000 & 0.343 & 1000 & 0.114 & 1000  \\ 
$\chi^2$ SVM   & {\bf 0.358}     &   1000    &  0.548    &  1000     & 0.375     &1000       & 0.338     & 1000      &  0.200    & 1000      \\ 
MKL-$\chi^2$  & 0.298     & 511       &0.554      &  62     &  0.381    & 472    & 0.316     &  195      & 0.199     & 471    \\ 
FS-linear &  0.239    & 350       & 0.535     & 422    & 0.315     & 665       & 0.355     & 384       & 0.186     & 443     \\ \hline
FS-$\chi^2$   &  0.304    & 62     & {\bf 0.565} & 75 & {\bf 0.384}& 284      &{\bf 0.366} &64         & {\bf 0.215}& 474      \\ \hline
\end{tabular}
\newline
\vspace*{0.1in}
\newline
\begin{tabular}{|l|c|r|c|r|c|r|c|r|c|r|}
\hline
\multirow{2}{*}{} & \multicolumn{2}{c|}{diningtable} & \multicolumn{2}{c|}{dog} & \multicolumn{2}{c|}{horse} & \multicolumn{2}{c|}{motorbike} & \multicolumn{2}{c|}{person} \\ \cline{2-11}
                   & AP & \#Feat & AP & \#Feat & AP & \#Feat & AP & \#Feat & AP & \#Feat \\ \hline
linear SVM   & 0.245 & 1000 & 0.278 & 1000 & 0.427 & 1000 & 0.289 & 1000 & 0.648 & 1000 \\ 
$\chi^2$ SVM  &{\bf 0.308} & 1000      & 0.337     & 1000    & {\bf 0.587}&1000       & 0.358     & 1000    &  0.689    &1000    \\ 
MKL-$\chi^2$  & 0.265     & 559       & 0.342     & 527       & 0.535     & 614       & 0.315     & 616    &0.726      & 231        \\ 
FS-linear  & 0.228     & 665       &  0.306    & 769       &  0.431    & 379       & 0.295     &  376    & 0.697     & 484      \\ \hline
FS-$\chi^2$  & 0.264     & 569       &{\bf 0.347}    & 423     & 0.525     & 78        &{\bf 0.378}& 82    & {\bf 0.741} &   63     \\ \hline
\end{tabular}
\newline
\vspace*{0.1in}
\newline
\begin{tabular}{|l|c|r|c|r|c|r|c|r|c|r|}
\hline
\multirow{2}{*}{} & \multicolumn{2}{c|}{pottedplant} & \multicolumn{2}{c|}{sheep} & \multicolumn{2}{c|}{sofa} & \multicolumn{2}{c|}{train} & \multicolumn{2}{c|}{TV monitor}  \\ \cline{2-11}
                   & AP & \#Feat & AP & \#Feat & AP & \#Feat & AP & \#Feat & AP & \#Feat \\ \hline
linear SVM  & 0.122 & 1000 & 0.235 & 1000 & 0.225 & 1000 & 0.449 & 1000 & 0.252 & 1000  \\ 
$\chi^2$ SVM  &{\bf 0.176}&1000       & 0.225     & 1000      &{\bf 0.272}&1000       &{\bf 0.566} & 1000     &   0.330   &  1000          \\ 
MKL-$\chi^2$   & 0.102     & 219       & 0.204     & 163       & 0.262     & 584       & 0.502     & 385       & 0.280     &  595         \\ 
FS-linear   & 0.113     & 420       & 0.226     & 282       & 0.243     & 596       & 0.420     &  525      & 0.294     & 372          \\ \hline
FS-$\chi^2$  &{\bf 0.176} & 526     &  {\bf 0.236} &  179    &  0.259    & 589       & 0.516     & 428       &{\bf 0.341} & 54      \\ \hline
\end{tabular}
\end{table*}

\subsection{Region Selection Experiments} \label{sec:exp_reg_sel}
As mentioned in Setion~\ref{sec:reg_sel}, region selection requires over-segmenting the images and videos first. For images, we used a hierarchical image segmentation to obtain superpixels~\cite{Arbelaez11}. For action localization on the MSR Action II, we followed \cite{ChenYC12} and used a regular voxel segmentation. For object localization on YTO dataset, we used the streaming hierarchical segmentation method of \cite{XuC12} to get supervoxels.

\begin{figure*}[!tb]
\centering
\includegraphics[width=1\linewidth]{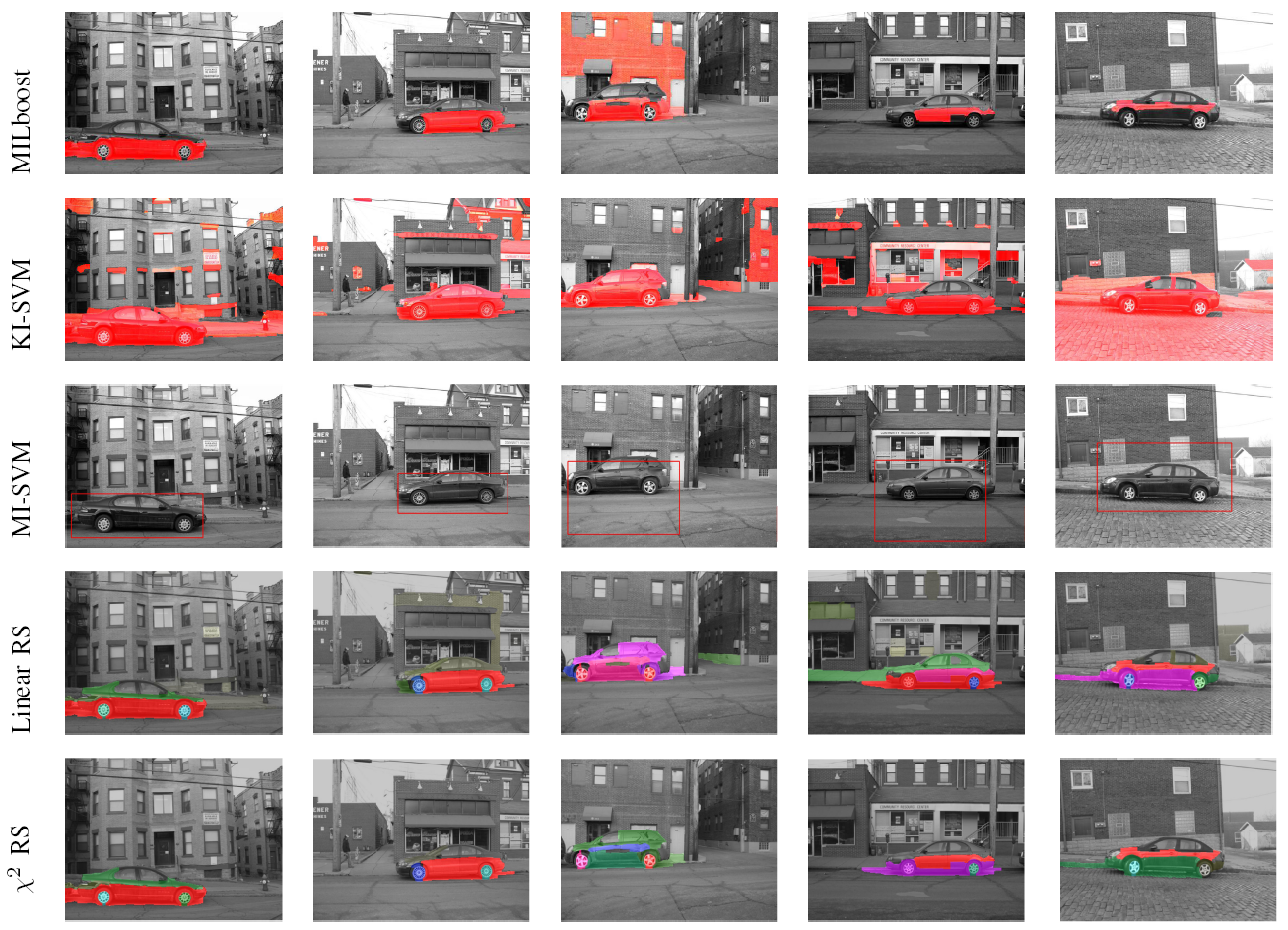}
\vspace{-0.1in}
\caption{Region selection for Pittsburgh Car dataset.  For our method (rows three and four) the color encodes the weights of the selected regions (warmer means higher); only regions with positive weights are colored. Images best seen in color.}
\label{fig:carlocalize}
\end{figure*}

{\bf PittCar:} Due to the connection of region selection to MIL approaches, we compared our region selection using linear and $\chi^2$ kernels with three popular MIL methods, MILboost~\cite{Viola05}, KI-SVM~\cite{LiYF09} and MI-SVM~\cite{Nguyen09}, on the PittCar dataset. We visualized the localization results in Fig.~\ref{fig:carlocalize}, from which we can see our region selection is visually best among these methods. In contrast, MILboost locates fewer regions, KI-SVM usually includes disperse background regions, and MI-SVM tends to include much background even though the size constraint has been imposed \cite{Nguyen09}.

\begin{figure}[!tb]
\centering
\subfigure[overlap threshold is $0.5$]
{\includegraphics[height=0.45\linewidth]{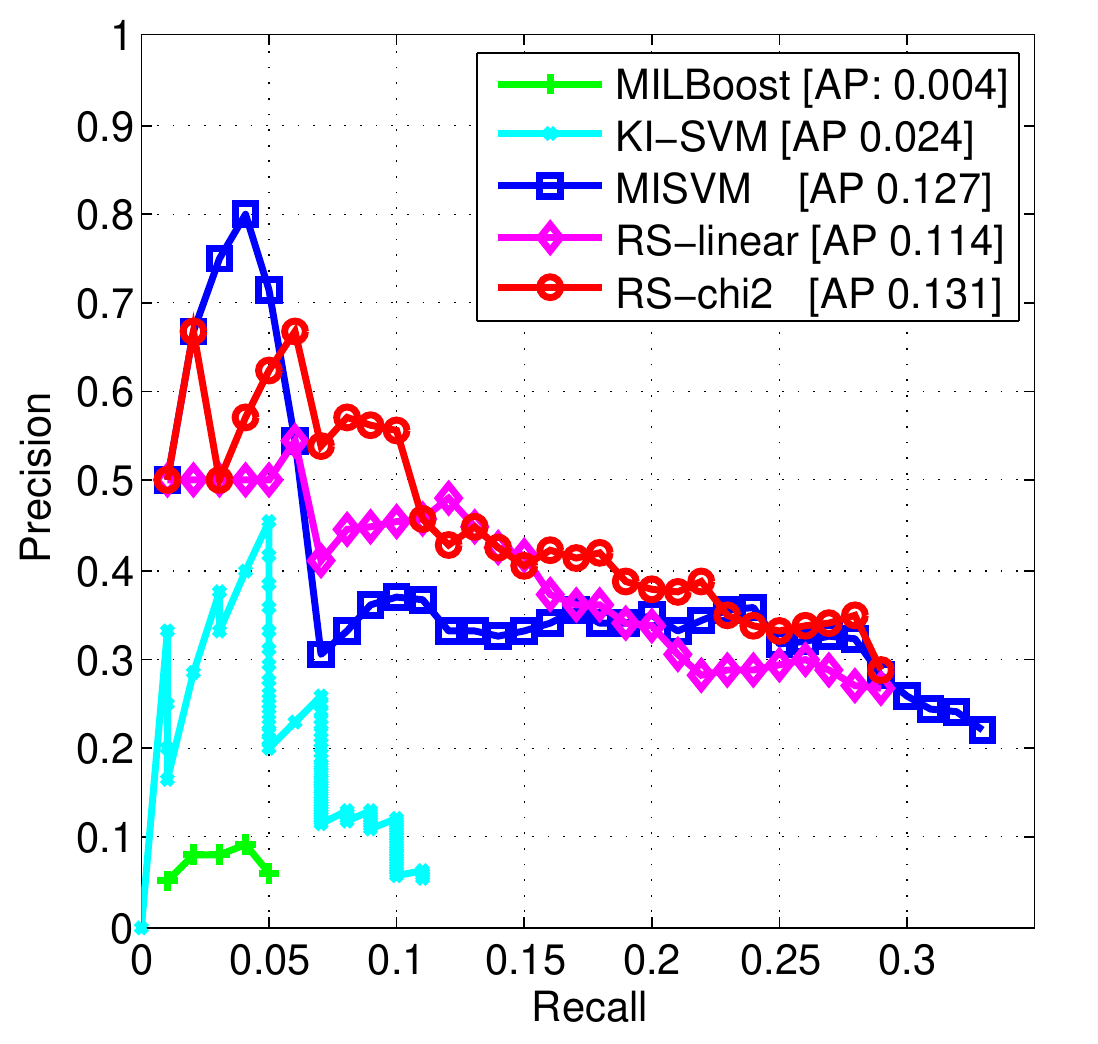}}
\subfigure[overlap threshold is $0.4$]
{\includegraphics[height=0.45\linewidth]{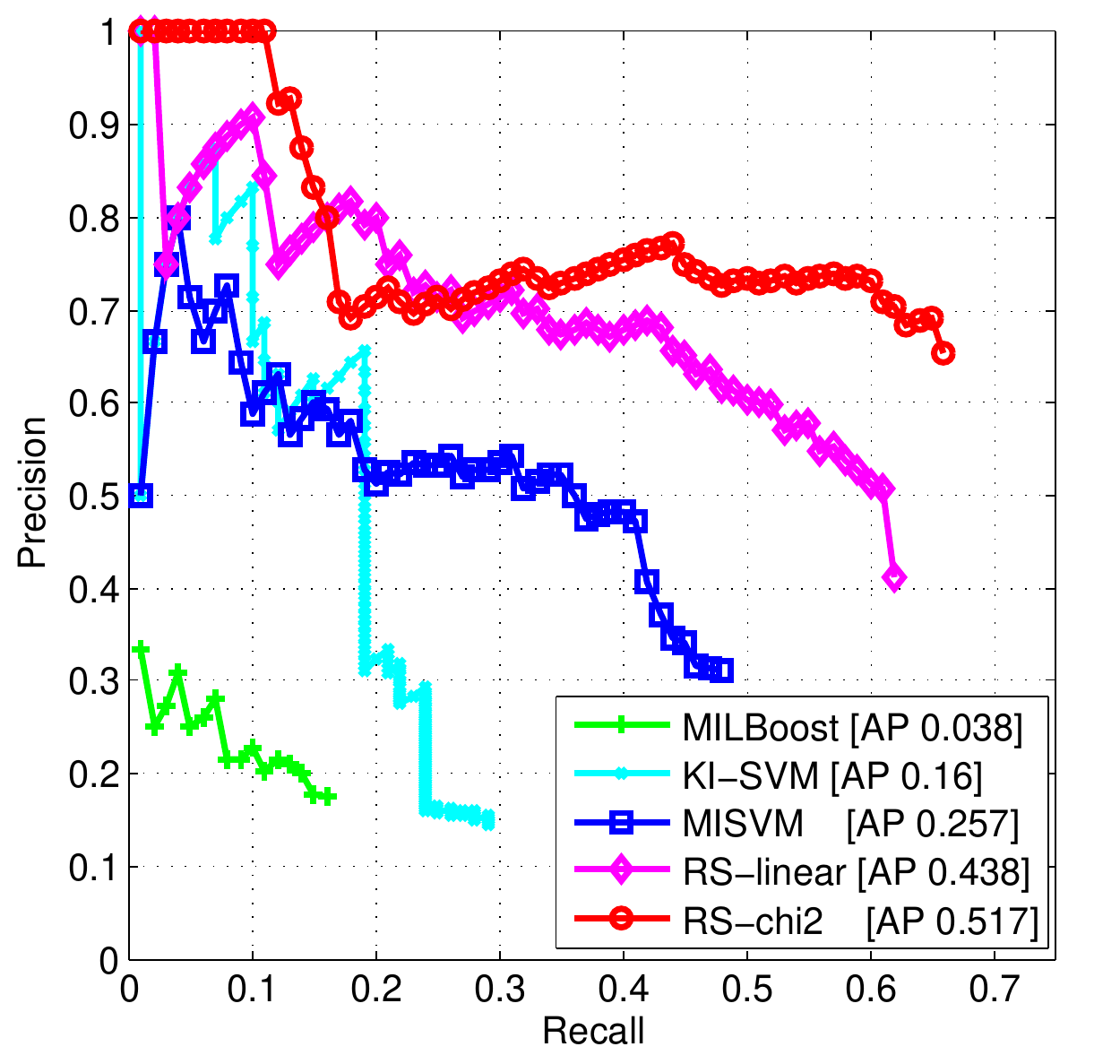}}
\vspace{-0.1in}
\caption{Localization performance on the PittCar dataset.}
\label{fig:prcar}
\end{figure}

To provide a quantitative measure for the localization performance, we compared all methods using precision-recall curves, as shown in Fig.~\ref{fig:prcar}. We used the area of overlap (AO) measure to evaluate the correctness of localization. For this criterion, a threshold $t$ should be defined for $AO$ to imply a correct detection. Usually, $t$ is set as $0.5$ \cite{Everingham09}. However, this is unfair for methods that localize arbitrary shape, because the ground truth is a bounding box and such methods provide a shape mask, which can yield more accurate localization. We thus also set $t$ to $0.4$. The PR curves of different $t$ values are shown in Fig.~\ref{fig:prcar}. We can see that our  method and MI-SVM perform comparably when $t=0.5$. For $t=0.4$, the region selection method performs significantly better than the baselines. Also, our region selection method using $\chi^2$ kernel performs better than with a linear kernel, which reinforces the usefulness of kernels in visual learning.

{\bf MSR Action II:} Since it is unclear how to apply the MI-SVM proposed in~\cite{Nguyen09} to video, we used the state-of-the-art method of Siva and Xiang~\cite{Siva11} as a baseline.

\begin{figure*}[!tb]
\centering
  {\includegraphics[width=1\linewidth]{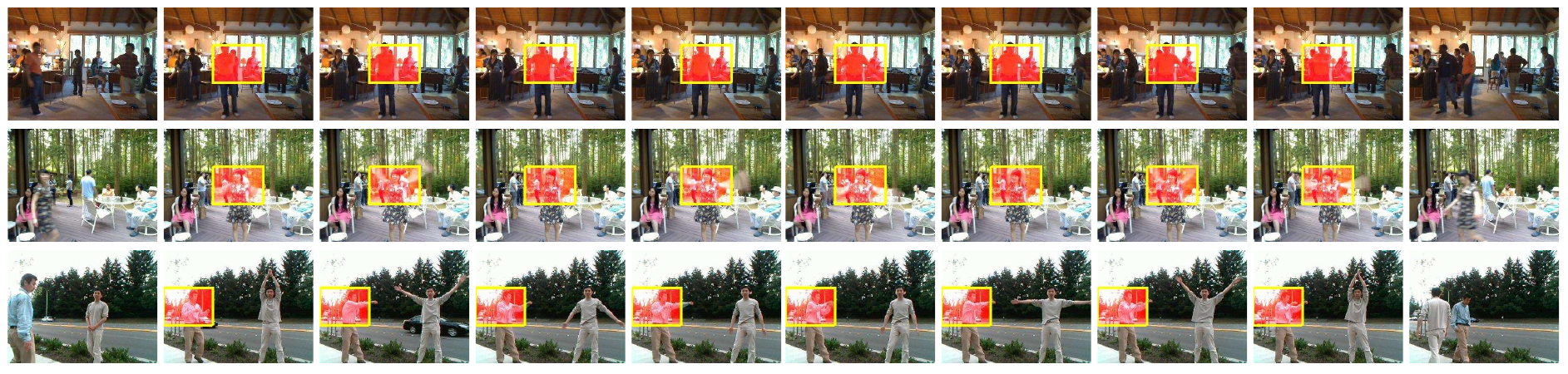}}
  \vspace{-0.2in}
\caption{Localization examples on MSR action II dataset. Each row corresponds to randomly selected $10$ frames in a video. Yellow bounding boxes are the localized actions in the videos.}
\label{fig:msr_det}
\end{figure*}

\begin{figure*}[!tb]
\centering
  {\includegraphics[width=0.32\linewidth]{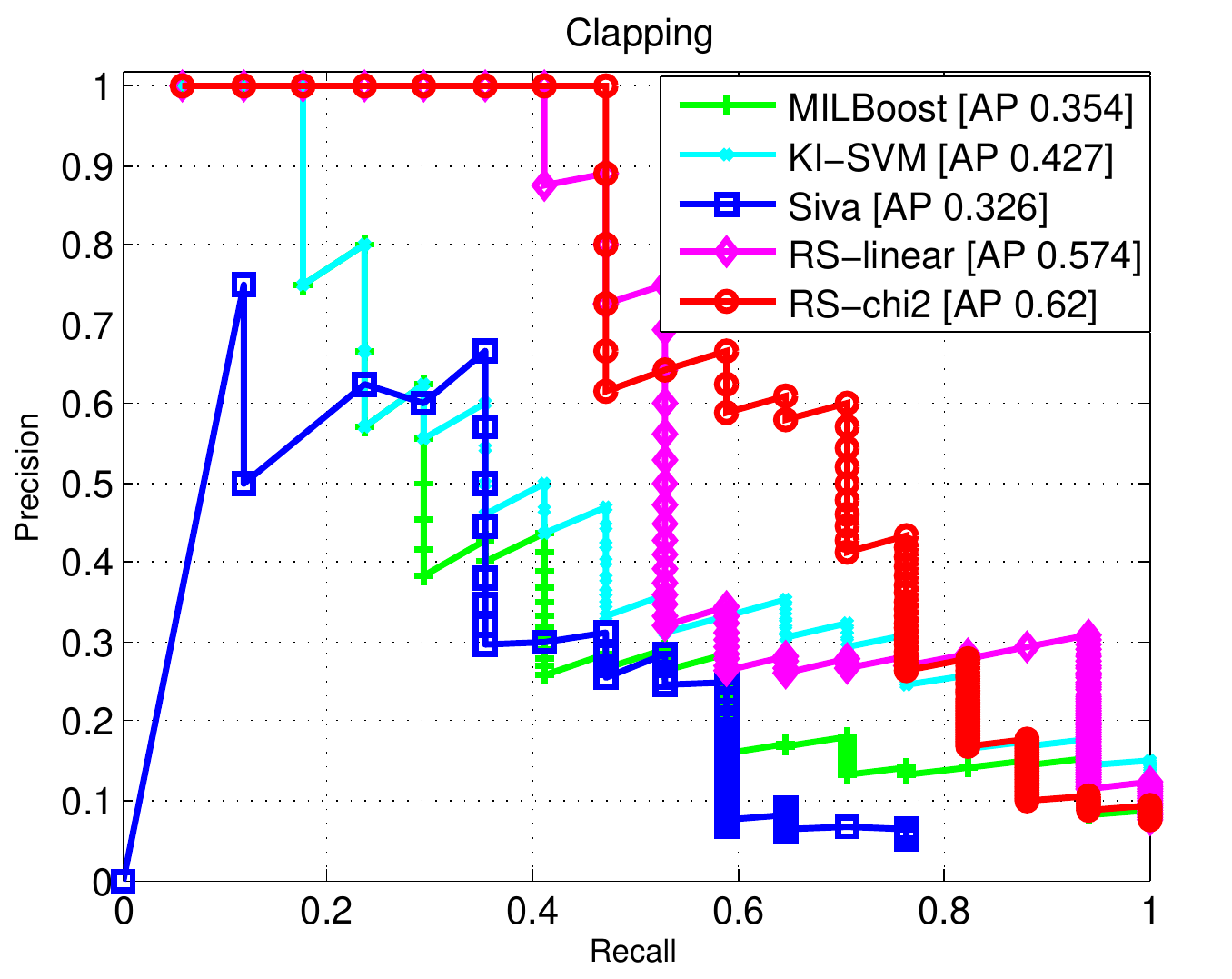}}
  {\includegraphics[width=0.32\linewidth]{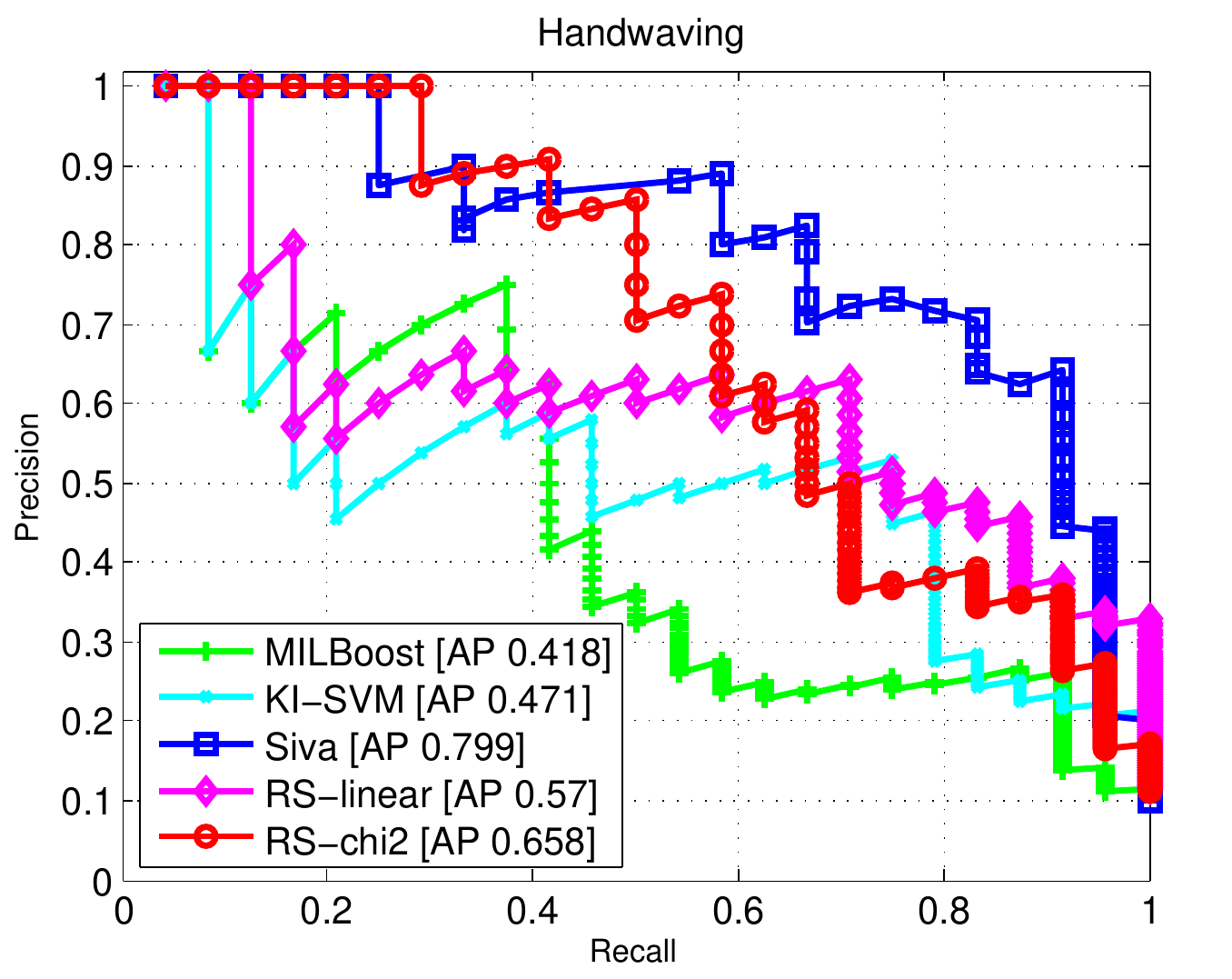}}
  {\includegraphics[width=0.32\linewidth]{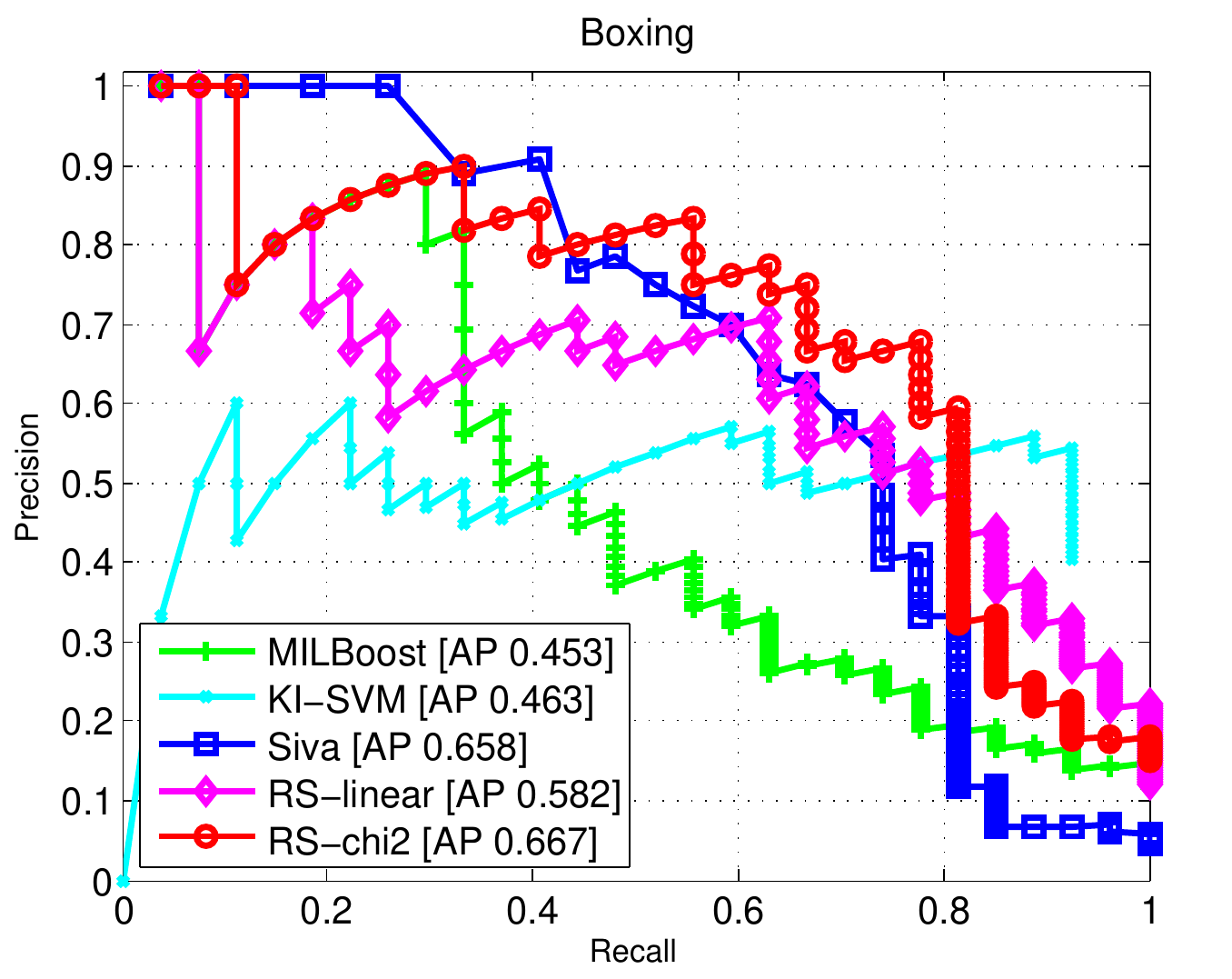}}
  \vspace{-0.1in}
\caption{Localization performance on MSR Action II.}
\label{fig:msrpr}
\end{figure*}

As in the previous experiment, we used precision-recall curve to evaluate the localization performance quantitatively. To ensure comparability, we replicate the setup of~\cite{Siva11} and set the temporal overlap to $1/8$ \cite{ChenYC12}. Qualitative and quantitative results are shown in Fig.~\ref{fig:msr_det} and Fig.~\ref{fig:msrpr} respectively. We can see that our region selection method using $\chi^2$ kernel (RS-chi2) performs better than linear kernel (RS-linear).  The region selection with a $\chi^2$ kernel outperforms both MILboost and KI-SVM significantly and yields comparable results to Siva and Xiang~\cite{Siva11}. Note, however,  that our method is independent of the video-segmentation methods, whilst the method of Siva explicitly assumes the use of human detector.


\begin{figure}[!t]
\centering
\begin{minipage}[c]{.6\textwidth}
  \centering
  {\includegraphics[width=1\linewidth]{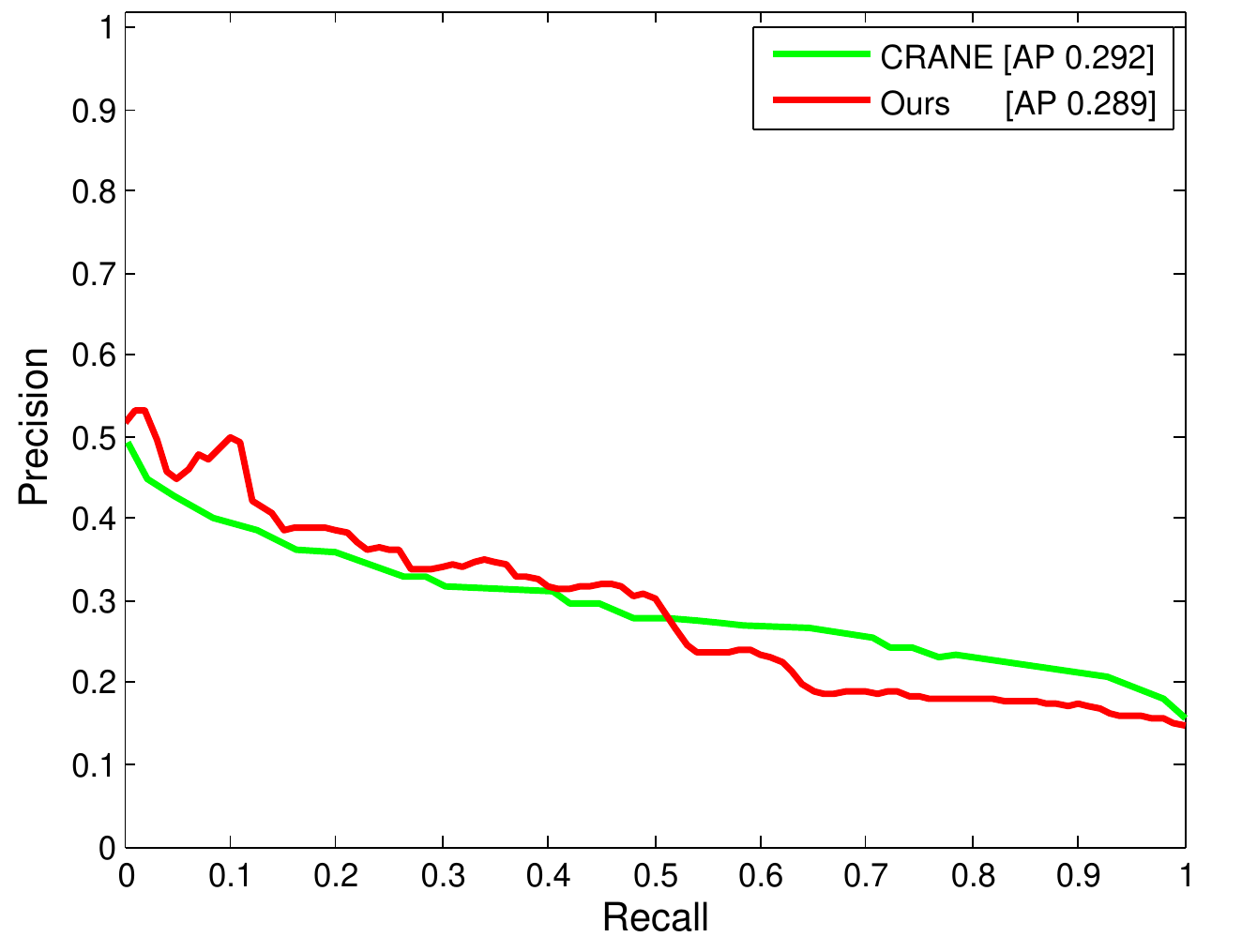}}
\end{minipage}
\begin{minipage}[c]{.3\textwidth}
  \centering
  \caption{Localization performance on YouTube-Objects dataset.}
  \label{fig:youtube}
\end{minipage}

\end{figure}

{\bf YouTube-Objects: } We also compared our region selection with CRANE~\cite{TangK13} which is the  state-of-the-art for object localization in videos. Here we use the $\chi^2$ kernel in our method. The average precision for each class is shown in Tab.~\ref{tab:yto}. We can see that our method gets better results on most of the vehicle categories and gets worse results on animal categories. The reason may lie in the pre-segmentation. Since animals are often small in these videos and perform non-rigid motion, the segmentation method we used can not provide as good segmentation as that used in \cite{TangK13}. In general, however, our result is comparable to CRANE, which can be seen from the averaged PR curve over classes in Fig.~\ref{fig:youtube}. However, it is important to note that our method
reported comparable results despite the fact that we used a worse segmentation algorithm.
\setlength{\tabcolsep}{4pt}
\begin{table*}[!t]
\begin{center}
\vspace{-0.1in}
\caption{Average precision on YouTube-Objects dataset.}
\label{tab:yto}
\begin{tabular}{|l|c|c|c|c|c|c|c|c|c|c|c|}
\hline
 & aero & bird & boat & car & cat & cow & dog & horse & mbike & train & \footnotesize{AVG.} \\
\hline
\footnotesize{CRANE} &  \footnotesize{0.365} & \footnotesize{\bf 0.363} & \footnotesize{\bf 0.271} & \footnotesize{0.446} & \footnotesize{\bf 0.250} & \footnotesize{\bf 0.334} & \footnotesize{\bf 0.345} & \footnotesize{\bf 0.286} & \footnotesize{0.158} & \footnotesize{0.204} & \footnotesize{\bf 0.292}\\
\hline
\footnotesize{Ours} &                \footnotesize{\bf 0.426} & \footnotesize{0.279} & \footnotesize{0.268} &       \footnotesize{\bf 0.612} & \footnotesize{0.204} & \footnotesize{0.203} & \footnotesize{0.283} & \footnotesize{0.148} & \footnotesize{\bf 0.202} & \footnotesize{\bf 0.263} & \footnotesize{0.289} \\
\hline
\end{tabular}
\end{center}
\end{table*}
\setlength{\tabcolsep}{1.4pt}

\section{Conclusions}
This paper proposes a feature and region selection method for visualization and understanding of the bag-of-words model.
These methods can also be used for image/video classification and weakly-supervised localization. A major advantage of our feature selection is that we can select features in the kernel space by solving a convex problem. This feature selection method achieves comparable accuracy to the state-of-the-art methods using significantly fewer number of features. In addition, our region selection method provides a tool to visualize the regions that the image/video classifier is weighting more aggressively to differentiate between class labels. The code is publicly available at \url{https://sites.google.com/site/drjizhao}.

While the method for feature selection is applicable to additive kernels, more research needs to be done to find convex solutions for non-additive kernels. In addition, other algorithms that can reduce the computational load of the optimization in space and time would be desirable. On the other hand, our region selection method can localize arbitrary shapes, beyond bounding-boxes; however, our method depends on the algorithm for over-segmentation and the object must be connected. These issues will remain to be explored in further research.

\appendix

\clearpage



\bibliographystyle{latex8}
\bibliography{manuscript}

\begin{thebibliography}{10}\setlength{\itemsep}{-1ex}\small

\bibitem{Andrews2002}
S.~Andrews, I.~Tsochantaridis, and T.~Hofmann.
\newblock Support vector machines for multiple-instance learning.
\newblock In {\em Advances in Neural Information Processing Systems 15}, pages
  577--584, MA, USA, 2002. MIT Press.

\bibitem{Arbelaez11}
P.~Arbel{\'a}ez, M.~Maire, C.~Fowlkes, and J.~Malik.
\newblock Contour detection and hierarchical image segmentation.
\newblock {\em IEEE Trans. Pattern Anal. Mach. Intell.}, 33(5):898--916, 2011.

\bibitem{Bilen14}
H.~Bilen, M.~Pedersoli, V.~P. Namboodiri, T.~Tuytelaars, and L.~V. Gool.
\newblock Object classification with adaptable regions.
\newblock In {\em Proc. IEEE Conf. Comput. Vis. Pattern Recognit.}, pages
  3662--3669, 2014.

\bibitem{Bradley98}
P.~S. Bradley and O.~L. Mangasaria.
\newblock Feature selection via concave minimization and support vector
  machines.
\newblock In {\em Proc. Int. Conf. Mach. Learn.}, pages 82--90, 1998.

\bibitem{CaoB07}
B.~Cao, D.~Shen, J.-T. Sun, Q.~Yang, and Z.~Chen.
\newblock Feature selection in a kernel space.
\newblock In {\em Proc. Int. Conf. Mach. Learn.}, pages 121--128, 2007.

\bibitem{Chatfield11}
K.~Chatfield, V.~Lempitsky, A.~Vedaldi, and A.~Zisserman.
\newblock The devil is in the details: an evaluation of recent feature encoding
  methods.
\newblock In {\em Proc. British Machine Vision Conference}, pages 76.1--76.12,
  2011.

\bibitem{ChenYC12}
C.-Y. Chen and K.~Grauman.
\newblock Efficient activity detection with max-subgraph search.
\newblock In {\em Proc. IEEE Conf. Comput. Vis. Pattern Recognit.}, pages
  1274--1281, 2012.

\bibitem{Do09}
H.~Do, A.~Kalousis, A.~Woznica, and M.~Hilario.
\newblock Margin and radius based multiple kernel learning.
\newblock In {\em Proc. Eur. Conf. Mach. Learn.}, pages 330--343, 2009.

\bibitem{Everingham09}
M.~Everingham, L.~V. Gool, C.~K.~I. Williams, J.~Winn, and A.~Zisserman.
\newblock The pascal visual object classes ({VOC}) challenge.
\newblock {\em Int. J. Comput. Vis.}, 88:303--338, 2009.

\bibitem{Faktor13}
A.~Faktor and M.~Irani.
\newblock Co-segmentation by composition.
\newblock In {\em Proc. Int. Conf. Comput. Vis.}, pages 1297--1304, 2013.

\bibitem{felzenszwalb10}
P.~F. Felzenszwalb, R.~B. Girshick, D.~McAllester, and D.~Ramanan.
\newblock Object detection with discriminatively trained part based models.
\newblock {\em IEEE Trans. Pattern Anal. Mach. Intell.}, 32(9):1627--1645,
  2010.

\bibitem{Feng15}
Y.~Feng and D.~P. Palomar.
\newblock Normalization of linear support vector machines.
\newblock {\em IEEE Trans. Signal Processing}, 63(17):4673--4688, 2015.

\bibitem{GaiK10}
K.~Gai, G.~Chen, and C.~Zhang.
\newblock Learning kernels with radiuses of minimum enclosing balls.
\newblock In {\em Advances in Neural Information Processing Systems 23}, pages
  649--657. Curran Associates, Inc., 2010.

\bibitem{Gehler08}
P.~V. Gehler and S.~Nowozin.
\newblock Infinite kernel learning.
\newblock Technical Report TR-178, Max Planck Institute for Biological
  Cybernetics, 2008.

\bibitem{Ghodrati14}
A.~Ghodrati, M.~Pedersoli, and T.~Tuytelaars.
\newblock Coupling video segmentation and action recognition.
\newblock In {\em Proc. IEEE Winter Conf. Applications of Comput. Vis.}, pages
  618--625, 2014.

\bibitem{Grandvalet02}
Y.~Grandvalet and S.~Canu.
\newblock Adaptive scaling for feature selection in {SVM}s.
\newblock In {\em Advances in Neural Information Processing Systems 15}, pages
  553--560, MA, USA, 2002. MIT Press.

\bibitem{Hartmann12}
G.~Hartmann, M.~Grundmann, J.~Hoffman, D.~Tsai, V.~Kwatra, O.~Madani,
  S.~Vijayanarasimhan, I.~Essa, J.~Rehg, and R.~Sukthankar.
\newblock Weakly supervised learning of object segmentations from web-scale
  video.
\newblock In {\em Proc. Eur. Conf. Comput. Vis. Workshop on Web-Scale Vision},
  pages 198--208, 2012.

\bibitem{Kloft11}
M.~Kloft, U.~Brefeld, S.~Sonnenburg, and A.~Zien.
\newblock $\ell_p$-norm multiple kernel learning.
\newblock {\em J. Mach. Learn. Res.}, 12:953--997, 2011.

\bibitem{Laptev05}
I.~Laptev.
\newblock On space-time interest points.
\newblock {\em Int. J. Comput. Vis.}, 64(2):107--123, 2005.

\bibitem{LiYF09}
Y.-F. Li, J.~T. Kwok, I.~W. Tsang, and Z.-H. Zhou.
\newblock A convex method for locating regions of interest with multi-instance
  learning.
\newblock In {\em Proc. European Conference on Machine Learning}, pages 15--30,
  2009.

\bibitem{LiuL12}
L.~Liu and L.~Wang.
\newblock What has my classifier learned? {V}isualizing the classification
  rules of bag-of-feature model by support region detection.
\newblock In {\em Proc. IEEE Conf. Comput. Vis. Pattern Recognit.}, pages 3586
  -- 3593, 2012.

\bibitem{Liu12}
X.~Liu, L.~Wang, J.~Yin, and L.~Liu.
\newblock Incorporation of radius-info can be simple with simple{MKL}.
\newblock {\em Neurocomputing}, 89:30--38, 2012.

\bibitem{lowe04}
D.~Lowe.
\newblock Distinctive image features from scale-invariant keypoints.
\newblock {\em Int. J. Comput. Vis.}, 60(2):91--110, 2004.

\bibitem{MaJ16}
J.~Ma, J.~Zhao, and A.~Y. Yuille.
\newblock Non-rigid point set registration by preserving global and local
  structures.
\newblock {\em IEEE Trans. Image Process.}, 25(1):53--64.

\bibitem{nguyen10}
M.~H. Nguyen and F.~{De la Torre}.
\newblock Optimal feature selection for support vector machines.
\newblock {\em Pattern Recognit.}, 43(3):584--591, 2010.

\bibitem{Nguyen09}
M.~H. Nguyen, L.~Torresani, F.~{De la Torre}, and C.~Rother.
\newblock Weakly supervised discriminative localization and classification: {A}
  joint learning process.
\newblock In {\em Proc. Int. Conf. Comput. Vis.}, pages 1925--1932, 2009.

\bibitem{Prest12}
A.~Prest, C.~Leistner, J.~Civera, C.~Schmid, and V.~Ferrari.
\newblock Learning object class detectors from weakly annotated video.
\newblock In {\em Proc. IEEE Conf. Comput. Vis. Pattern Recognit.}, pages
  3282--3289, 2012.

\bibitem{Rakotomamonjy08}
A.~Rakotomamonjy, F.~R. Bach, S.~Canu, and Y.~Grandvalet.
\newblock Simple{MKL}.
\newblock {\em J. Mach. Learn. Res.}, 9:2491--2521, 2008.

\bibitem{Raptis12}
M.~Raptis, I.~Kokkinos, and S.~Soatto.
\newblock Discovering discriminative action parts from mid-level video
  representations.
\newblock In {\em Proc. IEEE Conf. Comput. Vis. Pattern Recognit.}, pages
  1242--1249, 2012.

\bibitem{Russakovsky12}
O.~Russakovsky, Y.~Lin, K.~Yu, , and L.~Fei-Fei.
\newblock Object-centric spatial pooling for image classification.
\newblock In {\em Proc. Eur. Conf. Comput. Vis.}, pages 1--15, 2012.

\bibitem{Siva11}
P.~Siva and T.~Xiang.
\newblock Weakly supervised action detection.
\newblock In {\em Proc. British Machine Vision Conference}, pages 1--11, 2011.

\bibitem{TangK13}
K.~Tang, R.~Sukthankar, J.~Yagnik, and L.~Fei-Fei.
\newblock Discriminative segment annotation in weakly labeled video.
\newblock In {\em Proc. IEEE Conf. Comput. Vis. Pattern Recognit.}, pages 2483
  -- 2490, 2013.

\bibitem{Vedaldi12}
A.~Vedaldi and A.~Zisserman.
\newblock Efficient additive kernels via explicit feature maps.
\newblock {\em IEEE Trans. Pattern Anal. Mach. Intell.}, 34(3):480--492, 2012.

\bibitem{Vijayanarasimhan08}
S.~Vijayanarasimhan and K.~Grauman.
\newblock Keywords to visual categories: {M}ultiple-instance learning for
  weakly supervised object categorization.
\newblock In {\em Proc. IEEE Conf. Comput. Vis. Pattern Recognit.}, pages 1--8,
  2008.

\bibitem{Vijayanarasimhan11}
S.~Vijayanarasimhan and K.~Grauman.
\newblock Efficient region search for object detection.
\newblock In {\em Proc. IEEE Conf. Comput. Vis. Pattern Recognit.}, pages
  1401--1408, 2011.

\bibitem{Viola05}
P.~Viola, J.~C. Platt, and C.~Zhang.
\newblock Multiple instance boosting for object detection.
\newblock In {\em Advances in Neural Information Processing Systems 18}, pages
  1417--1424, MA, USA, 2005. MIT Press.

\bibitem{XuC12}
C.~Xu, C.~Xiong, and J.~J. Corso.
\newblock Streaming hierarchical video segmentation.
\newblock In {\em Proc. Eur. Conf. Comput. Vis.}, pages 626--639, 2012.

\bibitem{Yakhnenko11}
O.~Yakhnenko, J.~Verbeek, and C.~Schmid.
\newblock Region-based image classification with a latent {SVM} model.
\newblock Technical report, INRIA, 2011.

\bibitem{YangW12}
W.~Yang, Y.~Wang, A.~Vahdat, and G.~Mori.
\newblock Kernel latent {SVM} for visual recognition.
\newblock In {\em Advances in Neural Information Processing Systems 25}, pages
  818--826. Curran Associates, Inc., 2012.

\bibitem{YuanJ11}
J.~Yuan, Z.~Lin, and Y.~Wu.
\newblock Discriminative video pattern search for efficient action detection.
\newblock {\em IEEE Trans. Pattern Anal. Mach. Intell.}, 33(9):1728--1743,
  2011.

\bibitem{ZhuJ04}
J.~Zhu, S.~Rosset, T.~Hastie, and R.~Tibshirani.
\newblock 1-norm support vector machines.
\newblock In {\em Advances in Neural Information Processing Systems}, pages
  49--56, MA, USA, 2004. MIT Press.

\end{thebibliography}



\end{document}